\pdfoutput=1

\documentclass[11pt,a4paper,usenames,dvipsnames]{article}
\usepackage[hyperref]{eacl2021}
\usepackage[utf8]{inputenc}
\usepackage[T1]{fontenc}
\usepackage{times}
\usepackage{latexsym}
\usepackage{amsmath}
\usepackage{caption}
\usepackage{subcaption}

\usepackage{microtype}

\aclfinalcopy 
\def\aclpaperid{330} 

\usepackage{xspace}
\usepackage{booktabs}
\usepackage{graphicx}
\usepackage{enumitem}
\usepackage{xcolor}
\usepackage{tikz-dependency}
\usepackage{url}
\usepackage{amssymb}
\usepackage{float}

\setlist[description]{leftmargin=\parindent,labelindent=0pt}

\newcommand{\enquote}[1]{``#1''}

\newcommand{\sref}[1]{Sec.~\ref{sec:#1}}
\newcommand{\aref}[1]{Appendix~\ref{sec:#1}}

\newcommand{\tref}[1]{Table~\ref{tab:#1}}
\newcommand{\fref}[1]{Figure~\ref{fig:#1}}

\renewcommand{\vec}[1]{\mathbf{#1}}

\title{Coordinate Constructions in English Enhanced Universal Dependencies: Analysis and Computational Modeling}

\author{Stefan Grünewald$^{1, 2}$ \hfill Prisca Piccirilli$^{1, 2}$ \hfill Annemarie Friedrich$^2$\\	
	$^1$Institut für Maschinelle Sprachverarbeitung, University of Stuttgart\\
	$^2$Bosch Center for Artificial Intelligence, Renningen, Germany\\
	\texttt{stefan.gruenewald|annemarie.friedrich@de.bosch.com}\\
	\texttt{piccirpa@ims.uni-stuttgart.de}
}

\date{}

\begin{document}
\maketitle
\begin{abstract}
In this paper, we address the representation of coordinate constructions in Enhanced Universal Dependencies (UD), where relevant dependency links are propagated from conjunction heads to other conjuncts.
English treebanks for enhanced UD have been created from gold basic dependencies using a heuristic rule-based converter, which propagates only core arguments.
With the aim of determining which set of links \textit{should} be propagated from a semantic perspective, 
we create a large-scale dataset of manually edited syntax graphs.
We identify several systematic errors in the original data, and propose to also propagate adjuncts.
We observe high inter-annotator agreement for this semantic annotation task.
Using our new manually verified dataset, we perform the first principled comparison of rule-based and (partially novel) machine-learning based methods for conjunction propagation for English.
We show that learning propagation rules is more effective than hand-designing heuristic rules.
When using automatic parses, our neural graph-parser based edge predictor outperforms the currently predominant pipelines using a basic-layer tree parser plus converters.
\end{abstract}

\section{Introduction}
\label{section:intro}

The Universal Dependencies (UD) formalism \cite{de-marneffe-etal-2014-universal} is a framework for representing syntactic dependencies between words, prioritizing links between content words.
UD parses provide two levels of analysis.
\textit{Basic} dependencies form standard syntactic dependency trees in which each node has exactly one governor (black links on top in \fref{basicVsEnhanced}).
\textit{Enhanced} dependencies \citep{schuster-manning-2016-enhanced} are extensions of these trees including additional relations (blue links below sentence) with the aim of representing linguistic phenomena such as coordination, control, or relative clauses.
They have been shown to provide valuable input for information extraction tasks \citep{schuster2017paris}.
One of the most frequent phenomena addressed by enhanced UD is coordination.
In the English Web Treebank (EWT), more than 15\% of all sentences contain conjoined verbs.
Hence, a good representation of coordination clearly is crucial for downstream tasks.
For example, in \fref{basicVsEnhanced}, the enhanced layer explicitly captures that the arguments of  the predicate \enquote{wrote} also fill the corresponding slots of \enquote{published,} which is highly relevant for natural language understanding tasks.

\begin{figure}[t]
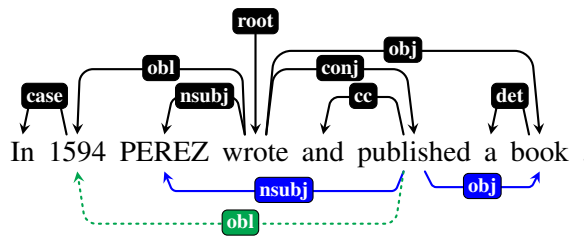

	\centering
	\hspace*{-4mm}
	\begin{dependency}[theme=night]
		\begin{deptext}
			\& In \& 1594 \& PEREZ \& wrote \& and \& published \&  a \& book \& .\\
		\end{deptext}
		\depedge{3}{2}{case}
		\depedge[edge unit distance=2.5ex,edge slant=1pt]{5}{3}{obl}
		\depedge{5}{4}{nsubj}
		\depedge[edge unit distance=2.5ex]{5}{7}{conj}
		\depedge{7}{6}{cc}
		\depedge{9}{8}{det}
		\deproot[edge unit distance=2.5ex]{5}{root}
		\depedge[edge unit distance=1.6ex]{5}{9}{obj}
		\depedge[edge below, edge style={blue}, label style={fill=blue}, edge unit distance=0.6ex]{7}{9}{obj}
		\depedge[edge below, edge style={blue}, label style={fill=blue}, edge unit distance=0.5ex]{7}{4}{nsubj}
		\depedge[edge below, edge style={dotted, Green}, label style={fill=Green}, edge unit distance=1ex]{7}{3}{obl}
	\end{dependency}
	\caption{UD \textbf{basic} (top) and \textcolor{blue}{\textbf{enhanced}} (bottom) dependencies. \textcolor{Green}{\textbf{Green dotted link:}} proposed addition.}
	\label{fig:basicVsEnhanced}
\end{figure}

In many cases, enhanced representations can be derived from the gold basic layer in a rule-based fashion \cite{schuster-manning-2016-enhanced}.
The currently available English enhanced UD treebanks have been created by applying such a converter.
However, we are not aware of a large study regarding their correctness and completeness.
Focusing on precision, the converter only propagates core arguments.
In this paper, we take a complementary approach, performing a large-scale annotation study in order to determine which set of links \textit{should} be propagated from a semantic perspective.
On a new dataset of 1,417 sentences from the EWT containing conjoined verbs, we verify and if necessary modify/extend the links involved in coordinate constructions.
We argue that adjuncts such as obliques should in fact be propagated at times, e.g., in \fref{basicVsEnhanced}, the additional (green dotted) link that we propose to add facilitates answering questions like \enquote{When was the book published?}.
To the best of our knowledge, our work constitutes the first large-scale annotation effort of this kind.

On the basis of our new dataset, we make the following contributions.
First, we estimate the degree of \textbf{correctness and completeness of the rule-based converter/existing treebanks}.
We find that the converter usually generates correct graphs when applied on gold basic trees, with some notable exceptions involving non-parallel syntactic constructions (e.g., conjuncts having different voice or mood).
In addition, the converter does not propagate links correctly in presence of multiple interacting conjunctions.
Our inter-anntotator agreement study shows high overlap for propagation decisions, with F1 between pairs of annotators of about 0.9 on average and around 0.75 for obliques.

Second, we address the question of \textbf{how to create high-quality treebanks} for enhanced UD from gold basic dependencies, again focusing on coordinate constructions.
Based on the findings of our corpus study, we improve the rule-based converter by \citet{schuster-manning-2016-enhanced}.
We also compare machine-learning (ML) based conjunction propagation classifiers in the form of (a) SVM-based classifiers as previously used for Finnish, Swedish and Italian \citep{nyblom-etal-2013-predicting,nivre-etal-2018-enhancing}, and (b) a novel neural approach integrating tree- and RoBERTa-based features.
We find that all systems mostly rely on tree-based features, but contextual embeddings also provide useful information.
Performance on propagation decisions has promising F1 around 0.9, already similar to human agreement.
ML-based classifiers outperform the rule-based converters on the EWT test set.

Third, we compare methods for \textbf{extracting propagated dependencies in an automatic parsing setting.}
The currently predominant approach is to run a basic-layer tree parser and then the same converter that has been used for gold standard construction.
We propose to use a neural graph-parser based edge predictor with an architecture similar to
\citet{dozat-manning-2018-simpler} instead, and show that this approach outperforms pipelines by around 9 points F1 on propagating links in conjunctions.
	
In sum, our contributions include: (1) a manually curated large-scale dataset of 1,417 sentences addressing semantically motivated correct and complete conjunction propagation in enhanced UD; (2) the proposal of novel neural approaches to conjunction propagation; and (3) experimental evidence that these models outperform rule- and pipeline-based approaches in both gold standard treebank enhancing and automatic parsing settings.
To the best of our knowledge, our work constitutes the first principled comparison of various approaches to propagating conjunctions in enhanced UD on manually corrected gold standard data for English.
Both our model implementations and the dataset are freely available.\footnote{\url{https://github.com/boschresearch/coordinate_constructions_english_enhanced_ud_eacl2021}}
We will contribute our changes to the EWT corpus to the next UD release.

\section{Related Work}
\label{sec:relwork}
\textbf{Coordinate Constructions in UD} are represented using the \textit{conj} relation, with the first conjunct being the head to which all dependencies of the phrase are attached (see \fref{basicVsEnhanced}).
In the basic layer, all governors and dependents of a conjoined phrase are attached to the first conjunct.
In the enhanced layer, relations are propagated to the dependent if suggested by the semantics of the sentence.\footnote{\url{https://universaldependencies.org/u/overview/enhanced-syntax.html}}
\newcite{schuster-manning-2016-enhanced} present an algorithm for creating enhanced dependencies automatically based on the basic layer.
While it propagates links with high precision, it propagates \textit{only} core arguments by design (see \aref{converter_details}).
In addition, it is highly reliant on correct basic dependencies (see \sref{experiments}).

\textbf{Conjunction propagation classifiers.}
\newcite{nyblom-etal-2013-predicting} present an SVM-based approach for enhancing Finnish syntax trees.
They observe high performance on conjunction propagation when operating on gold basic trees, but markedly worse results when using automatic parser output.
\newcite{nivre-etal-2018-enhancing} evaluate a similar approach for Swedish and Italian.
We show that their approach also works well for English, and extend it with neural models and contextualized word embeddings.
\citet{simi2018bootstrapping} de-lexicalize their rule-based converter developed for Italian, showing that their language-independent system also correctly produces most of the propagations for English.
However, they evaluate on EWT, which is itself the result of a rule-based system.
In contrast, we evaluate on manually checked gold data.

For the related task of dealing with \textbf{gapping} constructions such as \enquote{Paul likes coffee and Mary tea,} \newcite{schuster-etal-2018-sentences} reconstruct elided predicates by first parsing into an intermediate representation and then applying either a rule-based or an ML-based algorithm to copy over lexical material.
We here focus on dependency propagation and operate on gold tokens as annotated in the enhanced UD treebanks, which already include traces.
Other related work exists in the area of manual and rule-based error correction on UD treebanks \citep{wisniewski-2018-errator, alzetta-etal-2018-assessing}.

There is still little published work regarding \textbf{fully automatic enhanced UD parsing}, however, the topic has recently been addressed by the IWPT 2020 Shared Task \cite{EUDparsingST:2020}.
Among the top-performing systems, several approaches first parse into basic UD and then added transformation rules \citep[e.g.,][]{heinecke-2020-hybrid,dehouck-etal-2020-efficient}.
Others directly employ graph parsing techniques \citep[e.g.,][]{wang-etal-2020-enhanced,he-choi-2020-adaptation,hershcovich-etal-2020-kopsala}.
The overall winner TurkuNLP \citep{kanerva-etal-2020-turku} transforms enhanced UD into a tree format and then makes use of UDify \citep{kondratyuk-straka-2019-75}.
In addition, much work exists on \textbf{semantic dependency parsing} \citep[SDP,][]{oepen-etal-2014-semeval, oepen-etal-2015-semeval,may-priyadarshi-2017-semeval}.
These works differ from UD-based approaches as the respective formalisms represent meaning less close to syntactic structure, thus not requiring propagation.
From a modeling point of view, our work is most similar to that of \citet{grunewald-friedrich-2020-robertnlp}, who also use a graph-based biaffine architecture for enhanced UD parsing, and to that of \citet{dozat-manning-2018-simpler}, who achieve state-of-the-art results for SDP.

\section{Coordinate Constructions Dataset}
\label{sec:dataset}

In this section, we describe our creation of our manually created dataset and analyse the results.
	
\subsection{Data}
\label{sec:data}

\begin{table}[b]
	\centering
	\footnotesize
	\begin{tabular}{lrr}
		\toprule
		{}  &  \textbf{conj. sentences} &  \textbf{edited} \\
		\midrule
		\textbf{train}	& 1,926 	&		999	\\
		\textbf{dev}	 &		222		&		222	\\
		\textbf{test} 	&	196		&		196	\\
		\midrule
		\textbf{total} 	& 2,344   &  		1,417	\\
		\bottomrule
	\end{tabular}
	\caption{Coordinate constructions dataset statistics. \textbf{conj. sentences}: sentences in EWT containing verb phrase conjunctions; we edited 60\% of these.}
	\label{tab:numSent}
\end{table}

Our dataset consists of 1,417 sentences collected from EWT,\footnote{Linguistic Data Consortium  LDC2012T13.} containing data from five genres of web media (weblogs, newsgroups, emails, reviews, and Yahoo!\ answers).\footnote{\url{https://universaldependencies.org/treebanks/en_ewt/index.html}}
The basic dependencies of this UD gold standard have been derived from the original Stanford dependencies \citep{de-marneffe-etal-2006-generating} and were then hand-corrected.
The enhanced layer has been created using the automatic converter \citep[][see \aref{converter_details}]{schuster-manning-2016-enhanced}.
We retrieve all sentences containing at least one \textit{conj} link between two verbs.
More than 15\% of all sentences in EWT contain conjoined verbs.
Out of these sentences, we edit all sentences of the dev and test sets, and 999 sentences of the training set, amounting to more than 60\% of all relevant sentences in EWT (see \tref{numSent}).
The careful curation of each sentence took around 10 minutes on average, amounting to a total annotation effort of around 240 hours (total costs ca. \$4,750).
We exclude 18 sentences when reporting our statistics:
In 12 cases, the \textit{conj} relation is annotated wrongly in the basic layer and six sentences contain
syntactically non-standard English.\footnote{Such as \enquote{i want to be able to use it in my car, out n about etc...i guess like an iphone, but thats later on and ,i know what they are so no suggestions on just goin out to buy one im talking about right now just for an ipod??} (EWT dev set)}

\subsection{Annotation Methodology}
\label{sec:guidelines}
The manual corrections of the treebank were performed by a French native speaker with an extensive background in linguistics.
The annotation project involved regular discussions among all authors to decide on uncertain cases and to ensure consistency.
Additionally, in case of doubt, an English native speaker with an extensive linguistics background was consulted.
Dependencies were checked carefully sentence-wise using the ConLL-U-Editor tool \cite{heinecke-2019-conllueditor}.
If necessary, the full document was consulted to make sure interpretations were correct in context.

\paragraph{Annotation Guidelines.}

\begin{table}[t]
	\centering
	\footnotesize
	\begin{tabular}{lccc}
		\toprule
		& \textbf{A} & \textbf{B} & \textbf{C}\\
		\midrule
		\textbf{A} & - & 90.1 & 94.9\\
		\textbf{B} & 95.2 & - & 97.2\\
		\textbf{C} & 80.5 & 77.9 & - \\
		\bottomrule
	\end{tabular}
	\caption{Inter-annotator agreement on propagated links for 100 sentences: \textbf{precision} when treating the row annotator as gold standard (or, equivalently, \textbf{recall} when treating the column annotator as gold standard).}
	\label{tab:interAnnotator}
\end{table}

We verify and modify all links involved in coordinate constructions including conjoined verbs, but also noun or adjectival phrases.
First, we make sure that the automatically constructed enhanced representations adhere to the official guidelines for enhanced UD (see \sref{relwork}), propagating heads and dependents of conjuncts if the interpretation of the sentence suggests additional syntactic relations between words.

As each verb may also have its own complements, this task requires a semantic interpretation leveraging context and knowledge about selectional preferences.
If an ambiguity has already been resolved in the basic layer,\footnote{For example, in ``She was reading or watching a movie,'' ``movie'' is attached to the second conjunct ``watching'' in the basic layer, hence resolving the syntactic ambiguity.} we follow this interpretation unless obviously wrong.
Second, we propose to also propagate non-core dependents such as \textit{obl}, \textit{advcl} and \textit{advmod} if suggested by semantics, an annotation task similar to prepositional phrase attachment resolution.
We only propagate such links if the adjunct clearly modifies each conjunct (as in \fref{basicVsEnhanced}).
Finally, we extend the attachment of relative pronouns (\textit{ref}) to all antecedents if involved in coordinations.
We focus on propagating dependencies between content words, not propagating relations such as \textit{aux} or \textit{cop}, which could be handled as traces.

\paragraph{Inter-annotator agreement study.}
We sampled 100 sentences, half of them from cases where the primary annotator had judged the original version to be correct, and half of them cases that included modifications.
This sample was blindly re-annotated by two secondary annotators, both German native speakers with an extensive computational linguistics background.
\tref{interAnnotator} shows agreement in terms of precision and recall on the set of dependencies resulting from conjunction propagation, i.e., the links involved in conjunctions that are present in the enhanced layer but not in the basic layer.
For a formal definition, see \aref{agreement-scores}.
Agreement is generally high, particularly between annotators A and B.
Annotator C was more conservative in propagating links, especially in generally ambiguous cases.
However, the links that C propagates are also propagated by A and B.
Pairwise agreement was high on \textit{nsubj}, \textit{obj} and \textit{xcomp}.
Modifier clauses (\textit{acl}, \textit{advcl}) and adverbials (\textit{advmod}) were common sources of disagreement, indicating the more ambiguous nature of these propagations.
Pairwise scores and more details can be found in \aref{agreement-scores}.

\subsection{Analysis and Discussion}
\label{sec:data-analysis}

In this section, we analyse and discuss the modifications made to the original treebanks.

\paragraph{Quantitative Analysis of Changes.}
\label{sec:quant}
\tref{modification-stats} presents the numbers of dependency relations that have been added and removed in coordinate constructions in the enhanced layer.
More specifically, we consider only the set of links not present in the basic tree and count modifications regarding links starting or ending at conjuncts.\footnote{We made some additional fixes (not necessarily related to coordinations) to the original treebank, see \aref{full-modification-stats}.}
Counts for coarse-grained labels (e.g., \textit{nmod}) include all subtypes (e.g., \textit{nmod:for}) not explicitly listed in the table.
During our manual correction of the treebank, around 15\% of the total enhanced links involved in conjoined phrases were added and about 3\% were removed.
This confirms that the converter by \newcite{schuster-manning-2016-enhanced} is optimized for precision rather than recall, though our additions of course include labels that the converter does not address.
Note that in these cases, removed relations in \tref{modification-stats} are caused by fixes regarding attachment in the basic layer, whose errors had been propagated to the enhanced layer.
In total, we fixed errors in 57 sentences in the basic layer.
In 42 of these, this led to changes in the enhanced layer.

\begin{table}
	\footnotesize
	\centering
\begin{tabular}{lrrr|r}
	\toprule
	\textbf{label} &  \textbf{\#added} &  \textbf{\#removed} &  \textbf{\#sents} &  \textbf{\#total} \\
	\midrule 
acl &      14 &         4 &          12 &      68 \\
acl:relcl &      13 &         3 &           9 &     190 \\
advcl &      32 &         3 &          31 &     167 \\
advmod &      46 &         2 &          35 &       6 \\
amod &      19 &         2 &          14 &       7 \\
ccomp &       9 &        10 &          11 &      50 \\
nmod &      32 &         0 &          23 &     102 \\
nmod:poss &      11 &         0 &          10 &      18 \\
nsubj &     160 &        30 &         118 &     249 \\
nsubj:pass &       8 &        18 &          25 &     225 \\
nsubj:xsubj &      22 &        10 &          17 &    1688 \\
obj &       9 &         5 &          12 &      71 \\
obl &      72 &         0 &          51 &     150 \\
ref &      12 &         1 &           8 &      61 \\
xcomp &       7 &         5 &          10 &       8 \\
	\midrule
	\textbf{all} &    466 &        93 &         386 &    3060 \\
	\bottomrule
	\end{tabular}
	\caption{Statistics of modifications made to 1,399 sentences of the EWT.
	\textbf{\#sents} reports the number of sentences in which the respective reported changes were made, \textbf{\#total} reports the number of occurrences of the label in the enhanced layer of the original treebank.}
	\label{tab:modification-stats}
\end{table}

\paragraph{Linguistic Analysis of Changes.}
\begin{figure*}[t]
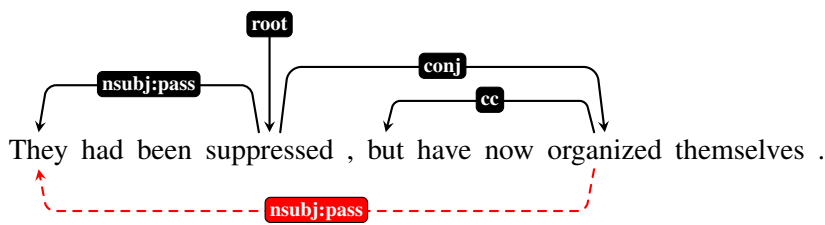
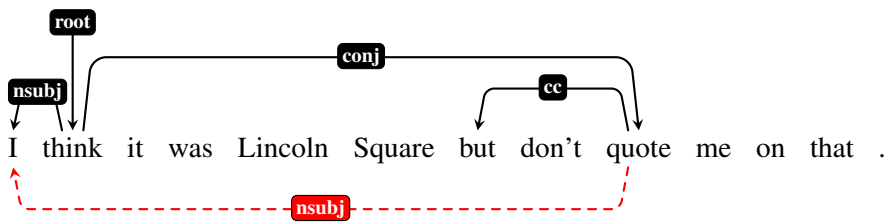
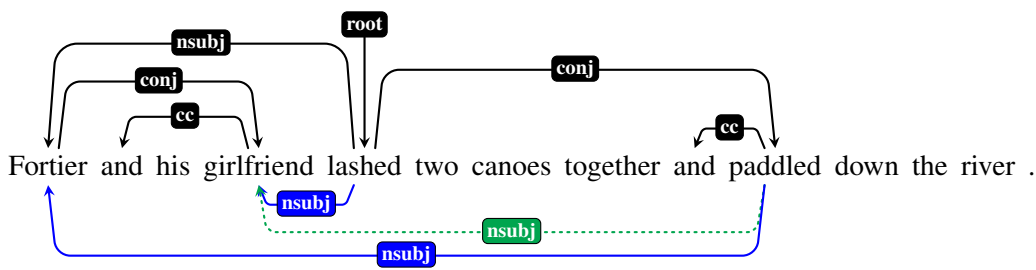

	\centering
	\begin{subfigure}[b]{0.98\textwidth}
		\centering
		\begin{dependency}[theme=night]
			\begin{deptext}
				They \& had \& been \& suppressed \& , \& but \& have \& now \& organized \& themselves \& .\\
			\end{deptext}
			\depedge[edge unit distance=1.2ex]{4}{1}{nsubj:pass}
			\deproot[edge unit distance=2.4ex]{4}{root}
			\depedge[edge unit distance=1.0ex]{4}{9}{conj}
			\depedge[edge unit distance=0.8ex]{9}{6}{cc}
			\depedge[edge below, edge style={dashed, red}, label style={fill=red}, edge unit distance=0.4ex]{9}{1}{nsubj:pass}
		\end{dependency}
		\caption{Passive voice: \textit{nsubj:pass} links should not be propagated verbatim if the second conjunct is in active voice. }
		\label{fig:error_passive}
	\end{subfigure}
	\begin{subfigure}[b]{0.98\textwidth}
		\centering
		\vspace{0.4cm}
		\begin{dependency}[theme=night]
			\begin{deptext}[column sep=0.15cm]
				I \& think \& it \& was \& Lincoln \& Square \&  but \& don't \& quote \& me \& on \& that \& .\\
			\end{deptext}
		    \depedge{2}{1}{nsubj}
			\deproot[edge unit distance=2.4ex]{2}{root}
			\depedge[edge unit distance=0.8ex]{2}{9}{conj}
			\depedge[edge unit distance=1.6ex]{9}{7}{cc}
			\depedge[edge below, edge style={dashed, red}, label style={fill=red}, edge unit distance=0.4ex]{9}{1}{nsubj}
		\end{dependency}
		\caption{Imperative mood: \textit{nsubj} links should not be propagated if the second conjunct is in imperative mood.}
		\label{fig:error_imperative}
	\end{subfigure}
	\begin{subfigure}[b]{0.98\textwidth}
		\centering
		\vspace{0.4cm}
		\begin{dependency}[theme=night]
			\begin{deptext}
				Fortier \& and \& his \& girlfriend \& lashed \& two \& canoes \& together \& and \& paddled \& down \& the \& river \& .\\
			\end{deptext}
			\depedge[edge unit distance=1.7ex]{1}{4}{conj}
			\depedge[edge unit distance=1.2ex]{4}{2}{cc}
			\depedge[edge unit distance=2.0ex]{5}{1}{nsubj}
			\deproot[edge unit distance=2.7ex]{5}{root}
			\depedge[edge unit distance=1.2ex]{5}{10}{conj}
			\depedge[edge unit distance=1.5ex]{10}{9}{cc}
			\depedge[edge below, edge style={blue}, label style={fill=blue}, edge unit distance=1.5ex]{5}{4}{nsubj}
			\depedge[edge below, edge style={dotted, Green}, label style={fill=Green}, edge unit distance=0.6ex]{10}{4}{nsubj}
			\depedge[edge below, edge style={blue}, label style={fill=blue}, edge unit distance=0.6ex]{10}{1}{nsubj}
		\end{dependency}
		\caption{Multiple coordinate constructions: \textit{nsubj} links should be propagated between the second conjuncts of each coordination.}
		\label{fig:error_multi}
	\end{subfigure}
	\caption{Systematic errors found in the automatic propagation of dependencies. (Only coordination-relevant links are depicted.) \textcolor{red}{\textbf{Red dashed link:}} Incorrect propagation or incorrectly labeled propagation. \textcolor{Green}{\textbf{Green dotted link:}} Missing propagation.}
	\label{fig:errorExamples}
\end{figure*}

One systematic error involves \textbf{links to subjects in passive constructions}: 18 out of 225 \textit{nsubj:pass} links were actually wrongly propagated.
All of them have been changed to \textit{nsubj}.
The reason is that the converter automatically propagates an \textit{nsubj:pass} link if the first conjoined verb is in the passive form, as, e.g., in 
\enquote{These Shiite movements had been suppressed by Saddam Hussein's regime, but have now organized and armed themselves} (see \fref{error_passive}).
Another common error (occurring 12 times) is the propagation of the first conjoined verb's subject to the second verb, even though the latter is in \textbf{imperative mood}, as, e.g., in \enquote{I think it was the Lincoln Square area but don't quote me on that} (see \fref{error_imperative}).

In sentences containing \textbf{multiple coordinate constructions}, such as \enquote{Dr. Fortier and his girlfriend lashed two canoes together and paddled eight kilometres along the Soper River,} \textit{nsubj} links should be present in the enhanced layer between both conjuncts of the subject noun phrase and both verbs.
However, in the original treebank, the second subject conjunct was never propagated to the second verb (see \fref{error_multi}).
Similarly, we also added many relations in cases of \textbf{nested coordinations} as in ``These Shiite movements had been suppressed by Saddam Hussein's regime, but have now organized and armed themselves.''
The second conjunct of the conjoined verb phrase is a conjoined verb phrase itself, but the \textit{nsubj} link to ``armed'' was missing.
In total, 194 sentences contain several coordinations, and we modified 92 of them.
This phenomenon also accounts for 45 of the added \textit{nsubj} links.

Some originally missing propagations concern \textbf{adjectival and adverbial modifiers} (\textit{acl, amod, advcl, advmod}), which are known to be ambiguous cases.
In ``Handwritten notes and files on a laptop were seized,'' the adjective ``handwritten'' clearly modifies the first conjunct ``notes'' only, but in ``Several Indian scholars and politicians have been ready to say and endorse anything,'' the propagation of ``several'' and ``Indian'' was added during our modifications.
These cases involve world knowledge that the converter currently does not handle. 

Finally, consider the sentence ``We recognize that the state may not require religious groups to officiate at, or bless, same-gender marriages.''
Both conjuncts take ``marriages'' as their argument, but as an \textit{obl} and as an \textit{obj} relation, respectively.
The resolution of such non-parallel constructions requires detailed subcategorization information.

\section{Modeling}
\label{sec:modeling}

In this section, we describe three approaches to generating links propagated due to coordination:
(1) an improved version of an existing converter (\sref{improved_converter}); (2) ML-based propagation classification operating on basic trees (\sref{conj-prop-classifiers}); and (3) a graph-parser based approach for directly predicting edges between tokens (\sref{graph-parser-model}).
While (1) and (2) may be used to construct ``silver standard'' enhanced UD graphs from gold trees, (3) is applicable in the automatic parsing setting only.

\subsection{Modifications to Rule-based Converter}
\label{sec:improved_converter}

Based on the error analysis in \sref{data-analysis}, we modify the rule-based converter by \newcite{schuster-manning-2016-enhanced} as follows.
In order to fix errors related to subject propagation in passive and imperative constructions, we take the conjunction dependent's morphological features into account.
In the gold standard, the \textit{Voice} feature is considered to be \textit{active} by default.
Hence, if the conjunction dependent does not have a \textit{Voice} feature or is explicitly marked as \textit{active}, an \textit{nsubj:pass} dependency will be propagated as \textit{nsubj}.
Similarly, if it has the feature \textit{Mood=Imp}, an \textit{nsubj} link will not be propagated.
Our second modification propagates common adjuncts of verbs as well (\textit{obl}, \textit{advmod}, and \textit{advcl}).
We maintain the rule from object propagation that a dependency is only propagated if the dependent comes after the potential target in the sentence.
Finally, to handle multiple and nested coordinations, we iterate the converter's conjunction propagation function until the dependency graph does not change any more.
This allows dependencies that result from propagation to be propagated themselves, retrieving links that would otherwise be missed.

\subsection{Conjunction Propagation Classifiers}
\label{sec:conj-prop-classifiers}

The core idea of ML-based conjunction propagation classifiers is to take a basic-layer tree and to decide for each incoming or outgoing dependency of the head of a coordinated phrase whether to propagate this dependency to the other coordinated item(s).
We refer to the coordinated nodes as \textbf{conjunction head} and \textbf{conjunction dependent} and to the candidate governor/dependent of the second conjunct as the propagation \textbf{target}.
In \fref{basicVsEnhanced}, these three nodes correspond to ``wrote,'' ``published'' and ``1954'' (or ``PEREZ''/``book''), respectively.
The output is a binary decision whether to propagate the given dependency or not.
In addition to the features described below, we always provide the candidate dependency label and direction.

\paragraph{SVM-based Classifier.}
We re-implement the method proposed by \citet{nyblom-etal-2013-predicting} using scikit-learn's SVC with a polynomial kernel of degree 2.\footnote{\url{https://scikit-learn.org}}
The features comprise morphological information about the tokens for the conjunction head/dependent and the target, as well as structural tree features extracted from the basic-layer tree.
For a detailed description, see \aref{features}.

\paragraph{Neural network classifier.}
\label{sec:bert_converter}
We pass the sentence through the transformer-based neural language model RoBERTa \citep{liu2019roberta} and extract the word embeddings for the first wordpiece tokens of the conjunction head, the conjunction dependent, and the propagation target.
In addition, we use equivalents of the SVM tree features using learned embeddings or one-hot encodings (see \aref{features}).
The inputs are concatenated and fed to a multi-layer perceptron, which then outputs the binary decision whether to propagate the dependency or not.
The multi-layer perceptron consists of two linear layers with hidden sizes 1500 and 500 respectively.
We implement the model using Huggingface's Transformers library \citep{Wolf2019HuggingFacesTS}.
RoBERTa weights are not fine-tuned.

\subsection{Graph-Parser Based Edge Prediction}
\label{sec:graph-parser-model}
In addition to the above approaches, we also evaluate a graph-parser based approach that predicts dependencies between tokens directly, i.e., which does not rely on a basic-layer tree.
Our unfactorized architecture is similar to that of \citet{grunewald-friedrich-2020-robertnlp}, i.e., our model predicts presence of edges and the corresponding labels in a single step, treating nonexistence of an edge as simply another label ($\varnothing$).
As we focus on the dependencies involved in conjunctions, we do not require the parser's output to constitute valid graphs.

Embeddings for input tokens are generated by feeding gold tokens to the RoBERTa tokenizer and then running the resulting word-pieces through the RoBERTa-large model.
We then generate an embedding $\vec{r}_i$ for the token at position $i$ by forming a weighted sum of the hidden layers' embeddings at the positions corresponding to the first word-piece token of the original token as suggested by \citet{kondratyuk-straka-2019-75}.
Weights for this scalar mixture of layers are learned during training.
Layers are randomly dropped during training to prevent the model from focusing on only a single layer.

For each input embedding $\vec{r}_i$, we create a head representation $h^{head}_i$ and a dependent representation $h^{dep}_i$ via two feed-forward neural networks:
\begin{align}
\vec{h}^{head}_i &= \text{FNN}^{head}(\vec{r}_i)\\
\vec{h}^{dep}_i &= \text{FNN}^{dep}(\vec{r}_i)
\end{align}

For each ordered pair $(i,j)$ of tokens, we feed their respective head and dependent representations to a biaffine classifier \cite{dozat2017deep} predicting logits $\vec{s}_{i,j}$ over the possible dependency labels. We use these logits to extract the probabilities $P(y_{i,j})$ for each label:
\begin{align}
\hspace*{-1mm}\text{Biaff}(\vec{x}_1, \vec{x}_2) &= \vec{x}^\top_1 \vec{U} \vec{x}_2 + W(\vec{x}_1 \oplus \vec{x}_2) + \vec{b} \label{eq:biaff}\\
\vec{s}_{i,j} &= \text{Biaff}\big( \vec{h}^{head}_i, \vec{h}^{dep}_j \big)\\
P(y_{i,j}) &= \text{softmax}(\vec{s}_{i,j})
\end{align}

$\vec{U}$, $W$ and $\vec{b}$ in (3) are learned parameters; $\oplus$ denotes concatenation.
The model is trained to minimize cross entropy loss w.\,r.\,t. the true dependency label between each pair of tokens.
If a token is not assigned any head due to $\varnothing$ scoring highest for all other tokens, we assign the highest-scoring non-$\varnothing$-relation and the corresponding head.

The model is simply trained to predict all link types in enhanced UD graphs.
In the training section of the EWT corpus, we replace every sentence that contains a coordinated verb phrase with our manually corrected version of that sentence, or remove it from the corpus if it is one of the 927 conjunction sentences in the training section which we did not correct.
For hyperparameter settings, see \aref{graph-parser}.

\section{Experiments}
\label{sec:experiments}

In this section, we describe our experiments on creating enhanced UD representations for coordinate constructions.
Analogous to \citet{nyblom-etal-2013-predicting}, we measure precision, recall and F1 on enhanced links that are the result of propagation in coordinate constructions.
For all experiments, we use gold sentence segmentation and tokenization, and evaluate on our manually corrected sentences from the dev and test sets of the EWT corpus.

\subsection{Gold Standard Treebank Enhancing}
\label{sec:exp-gold}

We first address the research question of how to best generate enhanced representations for treebanks with gold standard basic annotations.
We compare the following models: (1) an \enquote{Always} baseline, which simply propagates all incoming and outgoing links from the conjunction head to the conjunction dependent(s); (2) the rule-based converter by \citet{schuster-manning-2016-enhanced} and the variations thereof we developed inspired by our corpus study;
(3) our re-implementation of the SVM-based classifier by \citet{nyblom-etal-2013-predicting}; and (4) our neural-network (NN) based classifier.
The latter uses AdamW \citep{loshchilov2017decoupled} with a learning rate of 5e-5, a batch size of 1 and early stopping.
\tref{results_gold} reports the results on the development and test sets of our manually verified conjunction dataset.
The recall of the ``Always'' baseline is not at 100\% because a small number of relations change their label during propagation, e.g., \textit{nsubj}$\rightarrow$\textit{nsubj:pass}.

\paragraph{Rule-based conversion.}
We show results for successively adding components to the original converter (\textbf{RBC}).
On the test set, adding propagation of non-core dependents and allowing several iterations increases recall and improves F1 by more than 2 points.
On the dev set, in contrast, we do not observe these effects.\footnote{We assume that the reason is the presence of several informal-language sentences in the dev set that include multiple conjunctions (e.g., ``etc. etc. etc.''), whose annotation is unclear even in the basic gold standard.}
Adding our suggested passive/imperative fix surprisingly decreased performance.
Analysis showed that the cases that our converter got wrong were caused by erroneous morphological feature annotations in the basic layer.
In sum, our suggested improvements (\textbf{RBC2}) of heuristically propagating adjuncts (\textit{obl}, \textit{advmod}, \textit{acl}) and allowing several resolution passes of the converter seem to improve treebank enhancing, provided that the basic layer is correct.

\paragraph{ML-based conversion.}
Overall, the SVM and NN models
show similar performance.
As they perform already close to human agreement (see \tref{interAnnotator}), further improvement may actually indicate overfitting.
On the test set, the ML-based methods outperform the heuristic rule-based methods, surpassing the original converter by over 4 points F1.
We conclude that learning structural rules based on actual gold standard data is more effective than hand-designing them.
Differences on the dev set are less pronounced despite models being optimized on this data, again hinting to some qualitative differences between the two sets.

In order to determine which sources of information are most relevant, we perform ablation experiments for both classifiers.
The features representing the candidate dependency label and the direction of the link are essential and kept in each case.
Both the SVM and the NN classifiers draw most of their information from tree-based features.
This effect is particularly pronounced for the SVM classifier, where performance drops by 10 to almost 20 points F1 when ommitting these features.
The NN classifier's performance does not deteriorate as strongly under the same condition, indicating that some syntactic information can also be retrieved from contextualized word embeddings \citep[see e.g.,][]{tenney-etal-2019-bert}.
Nonetheless, in most experiments, adding token features improves performance slightly, showing that they do contain important information for propagation decisions.

\begin{table}[t]
	\centering
	\footnotesize
	\setlength\tabcolsep{3pt}
	\begin{tabular}{lccc|ccc}
		\toprule
		& \multicolumn{3}{c}{\textbf{Dev}} & \multicolumn{3}{c}{\textbf{Test}} \\
		& P & R & F & P & R & F\\
		\midrule
		\textit{``Always'' baseline} & \textit{23.1} & \textit{99.6} & \textit{37.5} & \textit{28.0} & \textit{99.6} & \textit{43.7}\\
		\midrule
		\textbf{RBC} & \textbf{94.8} & 86.4 & \textbf{90.4} & 95.2 & 76.9 & 85.0\\
		+ non-core deps & 93.7 & 86.4 & 89.9 & 94.9 & 79.7 & 86.7\\
		\hspace{2mm}+ iteration (\textbf{RBC2}) & 90.1 & 86.8 & 88.4 & 93.9 & 81.5 & 87.2\\
		\hspace*{4mm}+ passive fix & 91.7 & 85.3 & 88.4 & \textbf{95.7} & 78.6 & 86.3 \\
		\midrule
		\textbf{SVM} & 87.6 & 87.9 & 87.8 & 93.4 & 85.4 & \textbf{89.2}\\
		- tree features & 75.5 & 78.0 & 76.8 & 76.5 & 63.7 & 69.5 \\
		- token features & 86.3 & 87.5 & 86.9 & 92.3 & 85.1 & 88.5\\
		\midrule
		\textbf{NN} & 87.0 & 87.9 & 87.4 & 92.0 & \textbf{85.8} & 88.8\\
		- tree features & 87.1 & 86.4 & 86.8 & 88.0 & 78.6 & 83.1 \\
		- token features & 87.3 & \textbf{88.3} & 87.8 & 92.2 & 84.3 & 88.1\\
		\bottomrule
	\end{tabular}
	\caption{\textbf{Predicting relation propagation for coordinate constructions on gold basic trees}. Precision, recall and F1 on propagated relations in the predicted vs. gold dependency graphs in our manually verified conjunction dataset. The gold dev and test sets contain 273 and 281 instances, respectively.}
	\label{tab:results_gold}
\end{table}

\subsection{Propagating Conjunction Links in Automatic Parsing Setting}
\label{sec:exp-automatic}

For the scenario of parsing from raw tokens, we compare two state-of-the-art parsers, StanfordNLP \citep{qi-etal-2018-universal} and UDify \citep{kondratyuk-straka-2019-75}, combined with the rule-based converter or ML-based conjunction propagators, and our graph-parser based edge predictor.
The latter is trained on the subset of training sentences that either do not contain coordinated verb phrases or that were corrected by us.
Hyperparameters and training settings are given in \aref{graph-parser}.

Results for these experiments can be found in \tref{results_parsing-automatic}.
The impact of the quality of the parsed basic dependencies is evident: Results are much better for the UDify parser (LAS F1 of 89.4 for basic dependencies on the EWT dev set) than for StanfordNLP (LAS F1 of 87.4).
In the automatic setting, our heuristic extensions improve results compared to using the original converter, and there is no decrease in F1 on dev.
As in the gold standard settings, ML-based extensions improve upon RBC on test, but not dev.
Of the systems based on basic-layer tree parsers, RBC2 works best.
However, performance of all pipeline systems show rather poor performance at or below an F1 of 70.
Our graph-parser based edge predictor achieves by far the best results, outperforming all other models by a margin of over 7 points F1.
This shows that in an automatic setting, most robust results are achieved by directly inducing dependency links between tokens, modeling conjunction only indirectly.

To estimate the impact of our corrections to the gold standard, we also train the graph parser on uncorrected data.
The model trained on the corrected data has higher recall, but lower precision.
This is expected to some extent as we introduce semantically motivated propagations of adjuncts, and we suspect that they may require a larger training set.

\begin{table}[t]
	\centering
	\footnotesize
	\setlength\tabcolsep{3pt}
	\begin{tabular}{lccc|ccc}
		\toprule
		& \multicolumn{3}{c}{\textbf{Dev}} & \multicolumn{3}{c}{\textbf{Test}} \\
		& P & R & F & P & R & F\\
		\midrule
		Stanford+RBC & 70.8 & 63.0 & 66.7 & 56.5 & 47.7 & 51.7\\
		Stanford+RBC2 & 68.7 & 65.2 & 66.9 & 56.2 & 50.2 & 53.0\\
		Stanford+SVM & 64.3 & 65.2 & 64.7 & 54.7 & 49.8 & 52.1 \\
		Stanford+NN & 64.3 & 65.2 & 64.7 & 54.4 & 50.2 & 52.2 \\
		\midrule
		UDify+RBC & 72.8 & 67.8 & 70.2 & 71.8 & 58.0 & 64.2 \\
		UDify+RBC2 & 71.9 & 68.5 & 70.2 & 75.0 & 61.9 & 67.8 \\
		UDify+SVM & 70.6 & 68.5 & 69.5 & 70.9 & 59.1 & 64.5 \\
		UDify+NN & 69.9 & 68.9 & 69.4 & 70.4 & 60.1 & 64.9 \\
		\midrule
		GBP (orig. data) & \textbf{83.1} & 74.0 & 78.3 & \textbf{86.1} & 66.2 & 74.8\\
		GBP (our data) & 82.3 & \textbf{75.1} &  \textbf{78.5} & 82.5 & \textbf{68.7} & \textbf{75.0}\\
		\bottomrule
	\end{tabular}
	\caption{\textbf{Predicting relation propagation for coordinate constructions on parser output}. Otherwise same evaluation setup as in \tref{results_gold}.}
	\label{tab:results_parsing-automatic}
\end{table}

\subsection{Discussion}
The main insights comparing our experiments in the gold standard vs. the automatic parsing setting are as follows.
Overall, our heuristic extensions for the rule-based converter are beneficial in both settings.
In the gold setting, ML-based extensions lead to higher accuracy;  when applying them on noisy parser output, they do not work well.
However, using \textit{one} end-to-end machine-learning model directly to generate enhanced representations for conjunctions outperforms the pipeline version.
A possible reason for this might be that these models were all developed on gold data, while the graph-based parser does not rely on potentially wrong structural tree features and is also able to use internal confidence information for edges.
Another advantage of the end-to-end model may stem from the fact that its training allows to leverage semantic information from training data of a larger number of dependency links, i.e., including those not occurring in coordinate constructions.
This points to a promising future research direction, i.e, generating additional semi-artificial training data for conjunction propagation.

\section{Conclusion and Outlook}
We have presented a large-scale manually curated \textbf{dataset for conjunction propagation} in English.
In contrast to previous work focusing on high-precision rule-based propagation, we propagate links in all cases that semantically suggest argument or adjunct sharing.
In the gold standard treebank enhancing setting, we found \textbf{ML-based models} to outperform the de-facto standard rule-based converter by learning to exploit mostly structural features.
However, one of our main insights is that neither rule-based nor ML-based classifiers work well on noisy parser output precisely because of this reliance on structural information.
We propose to use a graph-parser based edge predictor instead and show that
it outperforms pipeline-based models by a large margin.
Our model reaches F1 scores between 0.75 and 0.78 with a precision of more than 0.82, a level of performance that may already be useful in downstream tasks.

Our models could be used for creating high-quality enhanced-level representations of conjunctions for the remaining English data, and could thus help in a UD community effort to continuously \textbf{improve the UD treebanks}.
Future work also includes the study of conjunction propagation methods for \textbf{further languages}.
Our in-depth study on English data provides several insights that we expect to be transferable cross-linguistically.
First, conjunction propagation can to some extent be addressed using heuristic rules, but capturing the full semantic nature of the task requires manual annotation.
Second, given appropriate training data, our machine-learning based approaches are also applicable to other languages.

In addition, it would be interesting to see if manually annotated data for coordinate constructions may be useful in natural language understanding tasks such as natural language inference (NLI). This is especially true for \enquote{stress test} datasets such as \textsc{ConjNLI} \cite{saha-etal-2020-conjnli}, which are designed to specifically test models' capabilities to process coordination.

Finally, as morphological features are generally important for this task, improving their automatic prediction \citep[see e.g.,][]{ramm-etal-2017-annotating,myers-palmer-2019-cleartac} as well as UD's gold standard seems to be a promising way to go.
Our work has demonstrated the value of a linguistically motivated corpus study of a syntactic-semantic phenomenon, and shown that given manually curated data, rules for conjunction propagation can be learned effectively.

\section*{Acknowledgments}
We would like to thank Sherry Tan and Johannes Hingerl for their support, as well as Jonas Kuhn, Heike Adel, Jannik Strötgen, and the anonymous reviewers for their useful comments regarding this work. In addition, we are grateful to the UD community, especially Nathan Schneider, for answering our questions.

\bibliography{references}
\bibliographystyle{acl_natbib}

\clearpage
\appendix

\section*{Appendix}
\label{sec:appendix}
\section{Detailed converter description}
\label{sec:converter_details}
In this section, we describe the algorithm implemented by the converter proposed by \citet{schuster-manning-2016-enhanced}.\footnote{\url{https://github.com/stanfordnlp/CoreNLP/blob/master/src/edu/stanford/nlp/trees/ud/UniversalEnhancer.java}}

For each \textit{conj} relation, the converter decides whether links ending or starting at the conjunction head (\texttt{gov}) should be propagated to the conjunction dependent (\texttt{dep}):

\begin{enumerate}
	\item Governors of \texttt{gov} are always propagated to \texttt{dep}, unless the relation is explicitly treated as an exception (e.g., \textit{vocative}, \textit{discourse}, or \textit{root}).
	\item Dependents of \texttt{gov} are propagated to \texttt{dep} as follows:
	\begin{enumerate}
		\item If the dependent is attached via \textit{nsubj} or \textit{csubj}, it is only propagated if \texttt{dep} does not already have a subject. If \texttt{dep} has an \textit{aux:pass} dependent, the relation is propagated as \textit{nsubj:pass} / \textit{csubj:pass}.
		\item If the dependent is attached via a non-subject core relation (\textit{obj}, \textit{iobj}, \textit{ccomp}, or \textit{xcomp}), it is propagated if and only if it comes after \texttt{dep} in the linear order of the sentence.
		\item Non-core dependents (such as \textit{obl}) are never propagated.
	\end{enumerate}
\end{enumerate}

The algorithm is able to handle many syntactically ambiguous cases, provided the underlying basic dependencies have resolved the ambiguity correctly.
Consider the sentence ``She was reading or watching a movie.''
If ``movie'' is correctly attached as an object of the conjunction dependent ``watching,'' it will not be propagated to ``reading'' in the enhanced representation.

\section{Inter-annotator agreement study}
\label{sec:agreement-scores}
Detailed comparisons between the three annotators can be found in \tref{agreementAvsB}, \tref{agreementAvsC}, and \tref{agreementBvsC}.
In our study, we consider only links that are part of the enhanced layer, but not of the basic layer.
For each annotator, we count for each label how often it occurs as an incoming or outgoing relation of a conjunct (columns labeled with the annotator's ID).
Formally, the set $E_A^l$ is the set of enhanced-layer edges that are (i) not present in the basic layer and (ii) involved in conjunctions as incoming or outgoing links of the conjuncts, with label $l$ marked by annotator $A$.
We also count the overlap of links for pairs of annotators.
Using these counts, we then compute precision, recall and F1, treating one annotator as the system and one as the gold standard.
For instance, when treating A as the gold standard and B as the system, this leads to:

\begin{align}
Precision_{BA} &= \frac{|E^l_A \cap E^l_B|}{E^l_B}\\
Recall_{BA} &= \frac{|E^l_A \cap E^l_B|}{E^l_A}
\end{align}

Note that when reversing this order, P and R are simply reversed, F1 stays the same.

The following numbers compare each annotator to the original gold standard (not in tables).
For modifier clauses (\textit{acl}, \textit{advcl}) and adverbials (\textit{advmod}), B was the most aggressive in propagating dependencies, adding 55 links in total for these labels, while A and C only added 39 and 32 links, respectively.
While all annotators propagated \textit{obl} dependencies roughly to the same extent, agreement was high between A and B but lower (F1 64-68\%) between C and the others, indicating that there are more ambiguities among these dependencies as well.
Annotator C is generally more conservative in propagating dependencies.
This is reflected in the relatively low recall when comparing to the other annotators, as well as the lower overall number of added links (285 as compared to 309 for A and 312 for B).

\vfill\break

\begin{table}[H]
	\centering
	\footnotesize
	\setlength\tabcolsep{5pt} 
	\begin{tabular}{lrrr|rrr}
		\toprule
		& \textbf{B} & \textbf{A} & \textbf{A\&B} & \textbf{P} & \textbf{R} & \textbf{F1} \\
		\midrule
		acl & 7 & 7 & 7 & 100.0 & 100.0 & 100.0 \\
		acl:relcl & 12 & 8 & 8 & 66.7 & 100.0 & 80.0 \\
		advcl & 24 & 17 & 17 & 70.8 & 100.0 & 82.9 \\
		advmod & 10 & 5 & 4 & 40.0 & 80.0 & 53.3 \\
		amod & 4 & 6 & 4 & 100.0 & 66.7 & 80.0 \\
		ccomp & 12 & 13 & 12 & 100.0 & 92.3 & 96.0 \\
		compound & 3 & 3 & 3 & 100.0 & 100.0 & 100.0 \\
		csubj & 2 & 2 & 2 & 100.0 & 100.0 & 100.0 \\
		nmod & 8 & 8 & 8 & 100.0 & 100.0 & 100.0 \\
		nsubj & 62 & 66 & 62 & 100.0 & 93.9 & 96.9 \\
		nsubj:pass & 5 & 5 & 5 & 100.0 & 100.0 & 100.0 \\
		nsubj:xsubj & 6 & 7 & 6 & 100.0 & 85.7 & 92.3 \\
		obj & 26 & 25 & 25 & 96.2 & 100.0 & 98.0 \\
		obl & 29 & 26 & 25 & 86.2 & 96.2 & 90.9 \\
		ref & 5 & 5 & 5 & 100.0 & 100.0 & 100.0 \\
		xcomp & 7 & 7 & 7 & 100.0 & 100.0 & 100.0 \\
		\midrule
		\textbf{total} & 222 & 210 & 200 & 90.1 & 95.2 & 92.6 \\
		\bottomrule
	\end{tabular}
	\caption{\textbf{Agreement} of Annotator A vs. Annotator B on links involved in coordinate constructions in the enhanced layer. For P/R computation, A was treated as the gold standard and B as the system.}
	\label{tab:agreementAvsB}
\end{table}

\begin{table}[H]
	\centering
	\footnotesize
	\setlength\tabcolsep{5pt} 
	\begin{tabular}{lrrr|rrr}
		\toprule
		& \textbf{C} & \textbf{A} & \textbf{A\&C} & \textbf{P} & \textbf{R} & \textbf{F1} \\
		\midrule
		acl & 7 & 7 & 7 & 100.0 & 100.0 & 100.0 \\
		acl:relcl & 4 & 8 & 4 & 100.0 & 50.0 & 66.7 \\
		advcl & 11 & 17 & 10 & 90.9 & 58.8 & 71.4 \\
		advmod & 4 & 5 & 0 & 0.0 & 0.0 & 0.0 \\
		amod & 4 & 6 & 4 & 100.0 & 66.7 & 80.0 \\
		ccomp & 11 & 13 & 11 & 100.0 & 84.6 & 91.7 \\
		compound & 3 & 3 & 3 & 100.0 & 100.0 & 100.0 \\
		csubj & 2 & 2 & 2 & 100.0 & 100.0 & 100.0 \\
		mark & 2 & 0 & 0 & 0.0 & 0.0 & 0.0 \\
		nmod & 6 & 8 & 6 & 100.0 & 75.0 & 85.7 \\
		nsubj & 61 & 66 & 61 & 100.0 & 92.4 & 96.1 \\
		nsubj:pass & 5 & 5 & 5 & 100.0 & 100.0 & 100.0 \\
		nsubj:xsubj & 6 & 7 & 6 & 100.0 & 85.7 & 92.3 \\
		obj & 25 & 25 & 23 & 92.0 & 92.0 & 92.0 \\
		obl & 18 & 26 & 18 & 100.0 & 69.2 & 81.8 \\
		ref & 2 & 5 & 2 & 100.0 & 40.0 & 57.1 \\
		xcomp & 7 & 7 & 7 & 100.0 & 100.0 & 100.0 \\
		\midrule
		\textbf{total} & 178 & 210 & 169 & 94.9 & 80.5 & 87.1 \\
		\bottomrule
	\end{tabular}
		\caption{\textbf{Agreement} of Annotator A vs. Annotator C on links involved in coordinate constructions in the enhanced layer. For P/R computation, A was treated as the gold standard and C as the system.}
	\label{tab:agreementAvsC}
\end{table}

\begin{table}[H]
	\centering
	\footnotesize
	\setlength\tabcolsep{5pt} 
	\begin{tabular}{lrrr|rrr}
		\toprule
		& \textbf{C} & \textbf{B} & \textbf{B\&C} & \textbf{P} & \textbf{R} & \textbf{F1} \\
		\midrule
		acl & 7 & 7 & 7 & 100.0 & 100.0 & 100.0 \\
		acl:relcl & 4 & 12 & 4 & 100.0 & 33.3 & 50.0 \\
		advcl & 11 & 24 & 11 & 100.0 & 45.8 & 62.9 \\
		advmod & 4 & 10 & 3 & 75.0 & 30.0 & 42.9 \\
		amod & 4 & 4 & 4 & 100.0 & 100.0 & 100.0 \\
		ccomp & 11 & 12 & 10 & 90.9 & 83.3 & 87.0 \\
		compound & 3 & 3 & 3 & 100.0 & 100.0 & 100.0 \\
		csubj & 2 & 2 & 2 & 100.0 & 100.0 & 100.0 \\
		mark & 2 & 0 & 0 & 0.0 & 0.0 & 0.0 \\
		nmod & 6 & 8 & 6 & 100.0 & 75.0 & 85.7 \\
		nsubj & 61 & 62 & 61 & 100.0 & 98.4 & 99.2 \\
		nsubj:pass & 5 & 5 & 5 & 100.0 & 100.0 & 100.0 \\
		nsubj:xsubj & 6 & 6 & 6 & 100.0 & 100.0 & 100.0 \\
		obj & 25 & 26 & 24 & 96.0 & 92.3 & 94.1 \\
		obl & 18 & 29 & 18 & 100.0 & 62.1 & 76.6 \\
		ref & 2 & 5 & 2 & 100.0 & 40.0 & 57.1 \\
		xcomp & 7 & 7 & 7 & 100.0 & 100.0 & 100.0 \\
		\midrule
		\textbf{total} & 178 & 222 & 173 & 97.2 & 77.9 & 86.5 \\
		\bottomrule
	\end{tabular}
		\caption{\textbf{Agreement} of Annotator B vs. Annotator C on links involved in coordinate constructions in the enhanced layer. For P/R computation, B was treated as the gold standard and C as the system.}
	\label{tab:agreementBvsC}
\end{table}

\hfill
\pagebreak

\section{Statistics on Treebank Modifications}
\label{sec:full-modification-stats}
As mentioned in \sref{quant}, we made changes to the original treebank in 1,417 sentences containing coordinate verb phrases.
While we focused on the dependency links starting or ending at conjuncts, we also fixed some additional errors that we spotted during this process.
\tref{modification-stats-all} gives statistics on these modifications (compare to \tref{modification-stats}, which includes only changes related to conjuncts).

\begin{table}[H]
	\footnotesize
	\centering
	\begin{tabular}{lrrr|r}
		\toprule
		\textbf{label} &  \textbf{\#added} &  \textbf{\#removed} &  \textbf{\#sents} &  \textbf{\#total} \\
	\midrule
	acl &      18 &         5 &          14 &     749 \\
	acl:relcl &      15 &         5 &           9 &    5,086 \\
	advcl &      39 &        11 &          36 &    2,055 \\
	advmod &      46 &         4 &          37 &    4,426 \\
	amod &      20 &         3 &          15 &    3,570 \\
	case &       2 &         1 &           2 &    6,415 \\
	ccomp &      12 &        16 &          13 &    1,003 \\
	det &       2 &         1 &           2 &    6,448 \\
	nmod &      32 &         2 &          24 &    3,279 \\
	nmod:poss &      12 &         1 &          10 &    7,485 \\
	nsubj &     194 &        41 &         144 &    7,815 \\
	nsubj:pass &       9 &        21 &          25 &    1,381 \\
	nsubj:xsubj &      35 &        20 &          31 &    8,758 \\
	nummod &       3 &         0 &           3 &     814 \\
	obj &      10 &        10 &          14 &    4,673 \\
	obl &      78 &         5 &          56 &    5,478 \\
	punct &       3 &         3 &           3 &    9,508 \\
	ref &      21 &         2 &          16 &     348 \\
	xcomp &      14 &         6 &          12 &     758 \\
	\midrule
	\textbf{all} &     565 &       157 &         466 &   80049 \\
	\bottomrule
\end{tabular}
	\caption{Statistics of modifications made to 1,417 sentences of the EWT, including both basic and enhanced layer. \textbf{\#sents} reports the number of sentences in which the respective reported changes were made, \textbf{\#total} reports the number of occurrences of the label in the original treebank.}
\label{tab:modification-stats-all}
\end{table}

\section{ML-based classifiers: features}
\label{sec:features}
\tref{ml_features} lists the features used in our SVM and NN models.
Token features are extracted for conjunction head, conjunction dependent, and propagation target each.
In addition to the listed features, we also experimented with including lemmas and POS tags, but did not find them to be useful in our ablation experiments. 

\begin{table*}
	\centering
	\footnotesize
	\begin{tabular}{lp{5cm}ll}
		\toprule
		Feature name & Description & SVM & NN\\ 
		\midrule
		\textbf{\textit{Instance features}} &&& \\
		dependency label & label of candidate link & one-hot & 50-dim. embedding\\
		incoming/outgoing & whether the dependency being propagated is an outgoing or incoming link at the conjunction head & one-hot & 50-dim. embedding\\
		midrule
		\textbf{\textit{Token features}} &&& \\
		morphological features & values of the \textit{Number}, \textit{Person}, \textit{VerbForm}, and \textit{Voice} features & one-hot & - \\
		contextualized word embeddings & word embeddings as generated by the RoBERTa-base model & - & 768-dim. embedding \\
		\midrule
		\textbf{\textit{Tree features}} &&&\\
		linear dependency direction & whether the linear direction of the candidate dependency is the same as for the dependency being propagated (both-left, both-right, or differing-directions) & one-hot & 50-dim. embedding\\
		existing dependency & whether the conjunction dependent already has a dependency of this type (only relevant for outgoing links) & one-hot & 50-dim. embedding\\
		outgoing dependencies (head) & set of outgoing dependencies of the conjunction head & one-hot & one-hot\\
		outgoing dependencies (dep) & set of outgoing dependencies of the conjunction dependent & one-hot & one-hot\\
		\# coord.-items & number of items in the coordination & one-hot & scalar\\
		\bottomrule
	\end{tabular}
	\caption{Description of feature sets used in ML-based conjunction propagation models.}
	\label{tab:ml_features}
\end{table*}

\vfill\break

\section{Graph-based edge predictor: Training Setup}
\label{sec:graph-parser}

\paragraph{Label lexicalization.}
At training time, we only use a limited label set of 56 labels where lexical material is replaced with placeholders, such as \textit{obl:[case]}. At prediction time, we retrieve the missing lexical material from the dependency graph in a rule-based fashion. In the simplest case, this means simply substituting the word form of the dependent of the required type (e.g., a \textit{case} relation). In conjunctions, the token in question may not have its own dependent of the correct type, instead \enquote{inheriting} if from its conjunction head. In that case, we retrieve the lexical material from the conjunction head's dependent.

\paragraph{Hyperparameters}
We perform only a minimal amount of hyperparameter tuning, mostly sticking with the values used by \newcite{kondratyuk-straka-2019-75}.
One notable exception is the training regime, where we found low batch size and the AdamW optimizer to yield the best results.
The full hyperparameter configuration can be found in \tref{parser_hyperparameters}.

\begin{table}[H]
	\centering
	\footnotesize
	\begin{tabular}{ll}
		\toprule
		\multicolumn{2}{c}{\textbf{RoBERTa embeddings}} \\
		Embeddings dimension & 1024\\
		Token mask probability & 0.15\\
		Layer dropout & 0.1\\
		Hidden dropout & 0.2\\
		Attention dropout & 0.2\\
		Output dropout & 0.5\\
		\multicolumn{2}{c}{\textbf{Biaffine classifier}}\\
		Hidden size & 1024\\
		Dropout & 0.33\\
		\multicolumn{2}{c}{\textbf{AdamW Optimizer}}\\
		Batch size & 5\\
		Learning rate & $5e^{-6}$\\
		$\beta_1$, $\beta_2$ & 0.9, 0.999\\
		Weight decay & 0.0\\
		\bottomrule
	\end{tabular}
	\caption{Hyperparameters for our graph-based parser.}
	\label{tab:parser_hyperparameters}
\end{table}



\end{document}